\def\year{2020}\relax
\documentclass[letterpaper]{article} % DO NOT CHANGE THIS
\usepackage{aaai20}  % DO NOT CHANGE THIS
\usepackage{times}  % DO NOT CHANGE THIS
\usepackage{helvet} % DO NOT CHANGE THIS
\usepackage{courier}  % DO NOT CHANGE THIS
\usepackage[hyphens]{url}  % DO NOT CHANGE THIS
\usepackage{graphicx} % DO NOT CHANGE THIS
\usepackage{graphicx}  % DO NOT CHANGE THIS
\usepackage{times}
\usepackage{epsfig}
\usepackage{graphicx}
\usepackage{amsmath}
\usepackage{amssymb}
\usepackage{multirow}
\usepackage{multicol}
\usepackage{tabularx}
\usepackage{booktabs}
\usepackage{mathtools}
\usepackage{subfigure}

\title{Relative Attributing Propagation: Interpreting the Comparative Contributions of Individual Units in Deep Neural Networks}

\author{Woo-Jeoung Nam\textsuperscript{\rm 1}, Shir Gur\textsuperscript{\rm 3}, Jaesik Choi\textsuperscript{\rm 5}, Lior Wolf\textsuperscript{\rm 3,4}, Seong-Whan Lee\textsuperscript{\rm 1,2}
\\\textsuperscript{\rm 1}Department of Computer and Radio Communications Engineering, Korea University
\\\textsuperscript{\rm 2}Department of Artificial Intelligence, Korea University
\\\textsuperscript{\rm 3}The School of Computer Science, Tel Aviv University
\\\textsuperscript{\rm 4}Facebook AI Research
\\\textsuperscript{\rm 5}Graduate School of Artificial Intelligence, KAIST\\
}
 
\begin{document}
\maketitle

\begin{abstract}
As Deep Neural Networks (DNNs) have demonstrated superhuman performance in a variety of fields, there is an increasing interest in understanding the complex internal mechanisms of DNNs. In this paper, we propose Relative Attributing Propagation (RAP), which decomposes the output predictions of DNNs with a new perspective of separating the relevant (positive) and irrelevant (negative) attributions according to the relative influence between the layers.
The relevance of each neuron is identified with respect to its degree of contribution, separated into positive and negative, while preserving the conservation rule. 
Considering the relevance assigned to neurons in terms of relative priority, RAP allows each neuron to be assigned with a bi-polar importance score concerning the output: from highly relevant to highly irrelevant. Therefore, our method makes it possible to interpret DNNs with much clearer and attentive visualizations of the separated attributions than the conventional explaining methods. 
To verify that the attributions propagated by RAP correctly account for each meaning, we utilize the evaluation metrics: (i) Outside-inside relevance ratio, (ii) Segmentation mIOU and (iii) Region perturbation. In all experiments and metrics, we present a sizable gap in comparison to the existing literature. Our source code is available in \url{https://github.com/wjNam/Relative_Attributing_Propagation}.
\end{abstract}

\section{Introduction}
Despite the impressive performance, the adoption of Deep Neural Networks (DNNs) is sometimes hindered by a transparency issue that arises from the complex internal structure of DNNs. Many studies have recently attempted to resolve the lack of transparency in DNNs. The attributing methods~\cite{bach2015pixel,montavon2017explaining,kindermans2017learning,shrikumar2017learning,montavon2018methods,lapuschkin-ncomm19,lundberg2017unified,sundararajan2017axiomatic} reveal the significant factors of the input in making decisions by assigning a relevance score to the input layer.
\begin{figure}
  \centering
  \includegraphics[scale=0.4]{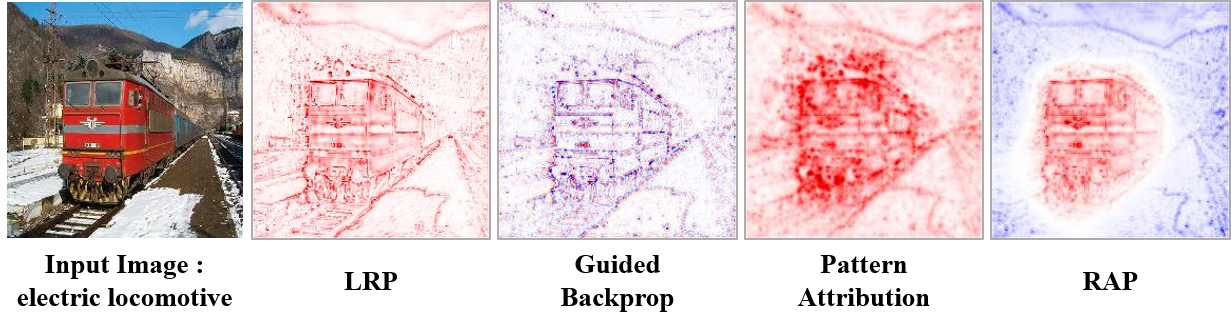}
    \caption{Comparison of the conventional explaining methods and RAP applied to VGG-16. In the previous methods, the attributions are similarly distributed across the entire image. Our RAP clearly distinguishes relevant (red) and irrelevant pixels (blue), placing the relevant attributions on the object, and the irrelevant ones on the background.}
  \label{fig:intro}
\end{figure}

To consider the positive and negative contributions of each image location to the output of a DNN, ~\cite{bach2015pixel} introduced the layer-wise relevance propagation rule, which propagates the relevance from the prediction. However, propagating the positive and negative relevance without considering the amount and direction of the contribution, may lead to defective interpretation. It is required to clarify the actual influences of individual units to the output, since the components of the complex inner structure shift and switch the conveyance of value. Furthermore, the relevance of each neuron is highly dependent on the absolute amount of contribution, resulting in both positive and negative relevance types to be correlated.

In this paper, we propose a new perspective for interpreting the relevance of each neuron, accounting for each neuron's influence among connected neighbors and allocating it with the relative importance. The main idea of this paper is changing the perspective of the relevance, from the sign of contribution to the influence among the neurons. Our method redistributes the relevance by changing the priority and rearranging it into positive and negative while preserving the conservation. This way, the relevance is assigned to each neuron directionally in line with the degree of importance to the output. 

Fig.~\ref{fig:intro} illustrates the comparison between RAP and conventional methods. While previous work considers directionality based on the sign of the neuron's contribution, which leads to a similar distribution of the positive and negative attributions, our method assigns the relevance according to the importance of neuron, which is highly focused on the object. 
The main contributions of this work are as follows:
\begin{itemize}
\item We propose relative attributing propagation (RAP), a method for attributing the positive and negative relevance to each neuron, according to its relative influence among the neurons. We address the phenomenon of the relevance dependency, which is highly dependant on the amount of neuron contribution and present the necessity of the new perspective to approach the relevance from the priority. We also prevent the risk of degeneracy during propagation by setting the criterion of separation according to the actual contribution between the intermediate layers.
\item 
We apply the Intersection of Union, Outside-Inside relevance ratio \cite{lapuschkin2016analyzing} and region perturbation \cite{samek2017evaluating} to assess whether the propagated attributions are meaningful. The evaluation shows that attributions from RAP provide a high objectness score with a clear separation of irrelevant regions, compared to the other explaining methods as well as to the literature of objectness methods.

\end{itemize}

%-------------------------------------------------------------------------

\section{Related Work}

There has been many recent studies on understanding of what a DNN model has learned. From the standpoint of interpreting a DNN model, the manner in which a DNN works can be visualized by maximizing the activation of hidden layers \cite{erhan2009visualizing} or generating salient feature maps \cite{dabkowski2017real,simonyan2013deep,mahendran2016visualizing,zhou2016learning,zeiler2014visualizing,zhou2018interpreting}. \cite{foerster2017input} introduced the input switched affine network, which can decompose the contributions of previous characters to the current prediction, and \cite{koh2017understanding} proposed the influence function to understand model behavior, debug models, detect dataset errors, and even create visually indistinguishable training-set attacks. \cite{ribeiro2016should} proposed LIME, an algorithm that explains the predictions of the classifier, by learning an interpretable model locally around the prediction.

From the standpoint of explaining the decision of a DNN, the contributions of the input are propagated backward, resulting in a redistribution of relevance in the pixel space. Sensitivity analysis visualizes the sensitivities of input images classified by a DNN while explaining the factors that reduce/increase the evidence for the predicted results \cite{baehrens2010explain}. \cite{zeiler2014visualizing} proposed a deconvolution method to identify the patterns of a predicted input image from a DNN. Layer-wise relevance propagation (LRP) \cite{bach2015pixel} was introduced to backpropagate the relevance, also called as attribution, by making the network output become fully redistributed throughout the layers of a DNN. \cite{samek2017evaluating} showed that the LRP algorithm qualitatively and quantitatively provides a better explanation than do either the sensitivity-based approach or the deconvolution method. 

Guided BackProp \cite{springenberg2014striving} and Integrated Gradients \cite{sundararajan2017axiomatic} each compute the single and average partial derivatives of the output to attribute the prediction of a DNN. Deep Taylor Decomposition \cite{montavon2017explaining} is an extension of LRP for interpreting the decision of a DNN, by decomposing the activation of a neuron in terms of the contributions from its inputs. DeepLIFT \cite{shrikumar2017learning} decomposes the output prediction by assigning the differences of contribution scores between the activation of each neuron to its reference activation. \cite{ancona2018towards} approached the problem of the attribution value from a theoretical perspective and formally proved the conditions of equivalence and approximation between four attribution methods: Guided Backprop \cite{springenberg2014striving}, Integrated Gradients \cite{sundararajan2017axiomatic}, LRP and DeepLIFT. \cite{lundberg2017unified} proposed SHAP explaining methods with unifying the conventional explaining methods and approximate the shapley value.

However, there are no studies which analyze the problem of ambiguous visualization in dealing with negative relevance. We bring out the fundamental causes of this problem and address the solution to handle the priority of neurons, resulting in the clear separation of the (ir)relevant objects. 

\section{Background}
\subsubsection{Notations}
Throughout this paper, the letter \(f(x)\) is used to denote the value of the network output before passing through the softmax layer for input \(x\). \(R\) represents the value of \(f(x)\) corresponding to the prediction class, which constitutes the input relevance for the attributing procedure. A neuron \(j\) in the layer \(l+1\) receives the value \(z_{ij}^{(l+1)}\) from a neuron \(i\) in the layer \(l\), which is obtained by multiplying the activation of the neuron in layer $l$, denoted \(m_i^{(l)}\), and the weight \(w_{ij}^{(l,l+1)}\). These contributions are summed over the the relevant neurons to obtain \(z_j^{(l+1)}\), which becomes \(m_j^{(l+1)}\) after adding the bias \(b_j^{(l,l+1)}\) and applying the activation function \(a(\cdot)\).

\begin{align}
    \begin{gathered}
        z_{ij}^{(l+1)} = m_i^{(l)}w_{ij}^{(l,l+1)},\,z_{j}^{(l+1)} = \sum_{i}z_{ij}^{(l,l+1)},\\
        m_{j}^{(l+1)} = a(z_{j}^{(l+1)} + b_{j}^{(l,l+1)})
    \end{gathered}
\end{align}

We consider the positive and negative parts of the contributions: $z_{ij} = z_{ij}^+ + z_{ij}^-$, where $z_{ij}^+ = max(z_{ij},0)$ and  $z_{ij}^- = min(z_{ij},0)$.
\subsubsection{Layerwise Relevance Propagation (LRP)}
LRP finds the parts with high relevance in the input, by propagating the relevance \(R\) from the output to the input.
The algorithm is based on the conservation principle, which maintains the relevance in each layer. Let \(R_{i}^{(l)}\) denote the relevance of a neuron \(i\) in a layer \(l\) and \(R_{j}^{(l+1)}\) is associated with a neuron \(j\) of the layer \(l+1\), this conservation takes the form:
\begin{equation}
    \sum_i R_{i}^{(l)} = \sum_j R_{j}^{(l+1)}
    \label{eq:was2}
\end{equation}
\cite{bach2015pixel} introduced two relevance propagation rules that satisfy Eq.~\ref{eq:was2}. The first rule, called LRP-\(\epsilon\), is defined as
\begin{equation}
    R_{i}^{(l)} = \sum_{j} \frac{z_{ij}}{\sum_{i}z_{ij}+\epsilon}R_{j}^{(l+1)}
    \label{eq:lrpeps}
\end{equation}
In this rule, a neuron \(i\) in the layer \(l\) receives the relevance, according to their contribution to the activation of the neurons in the layer \(l+1\). The constant \(\epsilon\) prevents the numerical instability for the case in which the denominator becomes zero. The second rule LRP-\(\alpha\beta\) enforces the conservation principle, while separating between the positive and negative activations in the relevance propagation process.
\begin{equation}
    R_{i}^{(l)} = \sum_{j} \left(\alpha\cdot\frac{z_{ij}^+}{\sum_{i}z_{ij}^+}-\beta\cdot\frac{z_{ij}^-}{\sum_{i}z_{ij}^-}\right)R_{j}^{(l+1)} \label{eq:lrp}
\end{equation}
Recall that \(z_{ij}^+ + z_{ij}^- = z_{ij}\). To maintain the total relevance, the parameters are chosen such that \(\alpha - \beta = 1\). We refer to the part of the relevance that is multiplied by $\alpha$ ($\beta$) and which is related to the positive (negative) activations as the positive (negative) relevance.

%%%%%%%%%%%%%%%%%
%%%%%%%%%%%%%%%%%

\section{The Shortcoming of Current Relevance Propagation Methods}
\begin{figure}
  \centering
  \includegraphics[scale=0.57]{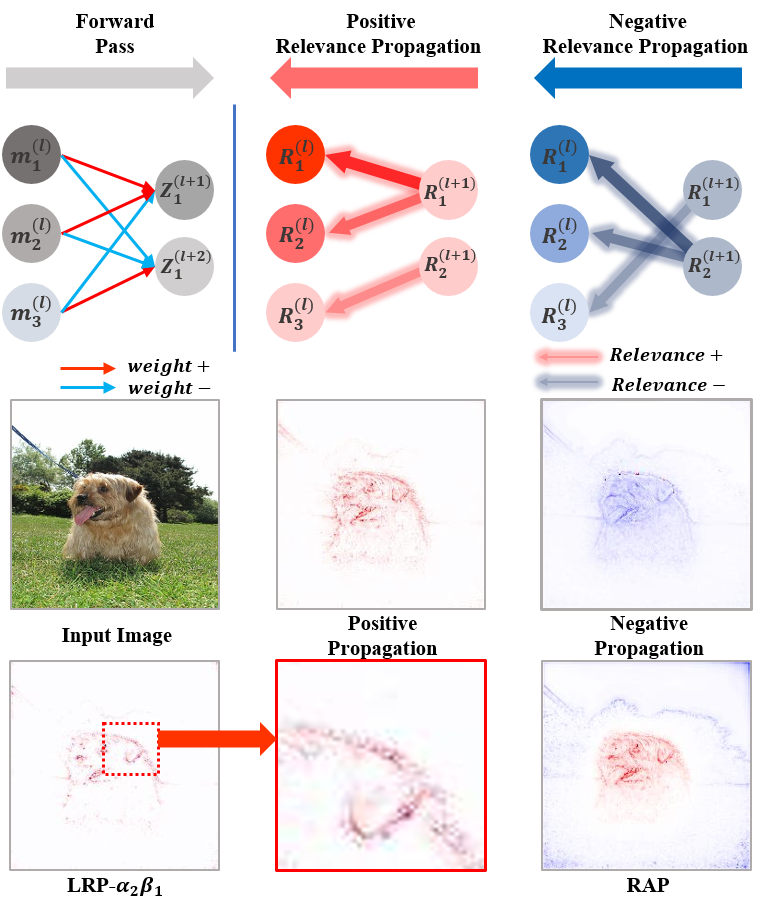}
  \caption{An illustration of the way positive and negative contributions are handled in LRP. See text for details.}
  \label{fig:was2}
\end{figure}
To motivate our method, we employ a toy sample to understand how relevance is propagated in the current methods. While it is presented in the context of $LRP_{\alpha\beta}$, similar pathologies can be found in other methods, such as $LRP_\epsilon$, integrated gradients, and pattern attribution. 
Fig.~\ref{fig:was2} presents an example of forward pass and backward relevance propagation between the two intermediate layers. For illustration purposes, the forward process does not include a bias term nor batch normalization, and all neurons in layer \(l\) are non-negative. For simplicity, the absolute values of all weights are identical. The darker color means the higher value of the neuron.
The positive part of \(R_{1}^{(l+1)}\) is propagated back to the neurons \(m_{1}^{(l)}\) and \(m_{2}^{(l)}\) at the ratio of \{\(m_{1}^{(l)}\):\(m_{2}^{(l)}\)\}. Similarly, the negative relevance of \(m_{1}^{(l)}\) receives in the propagation the lion's share of the relevance of \(R_{2}^{(l+1)}\), resulting in very low relevance value for \(R_{1}^{(l)}\). When summing the high positive and the negative contributions of neuron $i=1$ in Eq.~\ref{eq:lrp}, these cancel out. However, in terms of the amount of the contribution, it is clear that the relevance of this neuron should be high, since it plays a major role in the activations of layer $l+1$. 

Another related pathology is illustrated in the dog sample in Fig.~\ref{fig:was2}, which further illustrates this phenomenon. The relevance is propagated recursively using Eq.~\ref{eq:lrp} from the output layer to the input, obtaining the positive propagation image ($\alpha=1,\beta=0$). When doing the same process, but only for the negative value (similarly to $\alpha=0,\beta=1$), we obtain the negative propagation image. It seems that the same locations of the object receive both high positive relevance and high negative relevance. When these are combined, using Eq.~\ref{eq:lrp}, they tend to cancel each other. The combination of the positive and negative contributions is illustrated in the third image row, which demonstrated the results of the LRP-\(\alpha_{2}\beta_{1}\) method (similar to previous work this notation refers to the LRP method with $\alpha=2$ and $\beta=1$). Many of the positive relevance values are canceled out by equally large negative values, except for specific locations in which one contribution dominates. However, these locations can be either positive or negative and appear in close proximity to each other, as the zoomed-in subfigure demonstrates.

In RAP, we consider the absolute contribution of each neuron and propagating that relevance according to a novel method we introduce in the next section. The result of our method is shown on the bottom right of Fig~\ref{fig:was2}. As shown in the result, the positive and negative relevances appear in different parts of the image, corresponding to regions of high and low importance.

\section{Relative Attributing Propagation}
\begin{figure}[!t]
  \centering
  \includegraphics[scale=0.5]{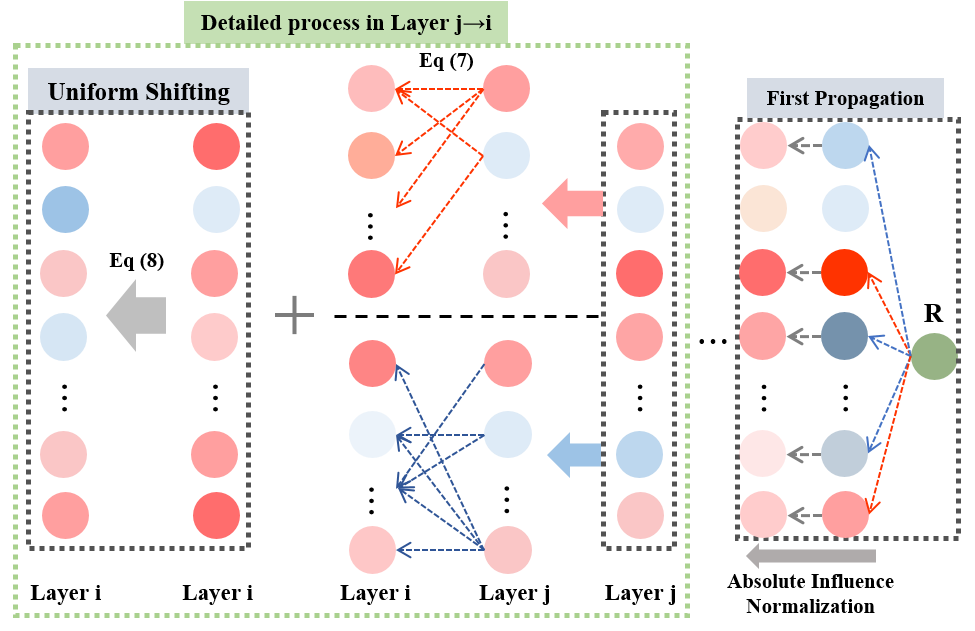}
  \caption{Overall structure of RAP algorithm.}
  \label{fig:rap}
\end{figure}

Motivated by the above issues, our goal is to separate the relatively (un)important neurons according to their influence across the layers. The method has three main steps: (i) absolute influence normalization, (ii) propagation of the relevance, and (iii) uniform shifting. Fig.~\ref{fig:rap} illustrates the three steps.
\subsubsection{Absolute Influence Normalization}
We first propagate the relevance value to neuron \(i\) of the penultimate layer \(p\), according to its actual contribution value \(m_{i}w_{ij}\) from the prediction node \(j\) in the final layer \(q\). Because the bias \(b_{j}^{(p,q)}\) is a single value, it is possible to consider the relevance of the bias \(b_{j}^{(p,q)}\) to the previous layer \(p\), by increasing the contribution of each neuron.
\begin{equation}
\begin{gathered}
    R_{i}^{(p)} = \left(\sum_{i}z_{ij}^++\sum_{i}z_{ij}^-\right) * \frac{R_{j}^{(q)} + b_{j}^{(p,q)}}{R_{j}^{(q)}}
\end{gathered}
\label{eq:was5}
\end{equation}
After applying the following, relevance values in the penultimate layer \(p\) are composed of both positives and negatives. 

Next, we normalize the entire value \(R_{i}^{(p)}\) by the ratio of the absolute positive and negative values \(|R_{i}^{(p)+}|:|R_{i}^{(p)-}|\).
\begin{equation}
R_{i}^{\prime(p)} = |R_{i}^{(p)}| * \frac{\sum_{i}R_{i}^{(p)}}{\sum_{i}|R_{i}^{(p)}|}
\label{eq:was6}
\end{equation}

\(R_{i}^{\prime(p)}\) is the new input relevance to the next propagation and all relevance values in layer \(p\) are distributed by their relative importance to the output layer, from most influenced to rarely influenced. The greater the neuron's influence on the contribution, the more positive relevance it is assigned. Eq.~\ref{eq:was6} is only applied in the first relevance propagation process.

\subsubsection{Criterion of Relevant Neuron}
After changing the relevance to the amount of absolute contribution each neuron has, i.e., when all relevance scores are positive, it is possible to propagate the relevance while maintaining a degree of relative influence. We then apply a uniform shifting to all activated neurons, which causes low influential neurons to have negative relevance. 

Next, the relevance propagation redistributes the relevance to the next layer through the positive contributions \(\mathcal{P} = \{i,j| z_{ij}^+>0\}\) and \(\mathcal{N} = \{i,j| z_{ij}^-<0\}\). It is possible to propagate the relevance of layer \(l\) through the formal case \(\mathcal{P}\), which makes each relevance to be redistributed according to the degree of the positive influence. For the negative case, we apply the same procedure, considering the influence among the negatively contributed neuron. Here, the propagated relevance originally means the ratio between the whole negative contribution in the forward pass. We utilize this amount of the relevance to uniformly shift the whole activated neurons, which makes the neurons to be converted as negative, in the order in which they are close to zero. For each relevance value in the layer \(l\), it is possible to compute the ratio of the positive and negative contribution with \(\{\sum_{j}z_{ij}^+:\sum_{j}z_{ij}^-\}\). However, when we compute the ratio with the original sign, the degeneracy problem has occurred because the absolute value of the numerator $\sum_{j}z_{ij}^+, \sum_{j}z_{ij}^-$ has a much larger value than the denominator $\sum_{j}z_{ij}^++\sum_{j}z_{ij}^-$. Therefore, the contribution ratio is computed after normalizing with the absolute value of each contribution. 
\begin{equation}
\begin{gathered}
% \bar R_{j}^{(l+1)}=
% \begin{cases}
% \rho_{j}^{(l+1)} = R_{j}^{(l+1)} * \frac{\sum_{i}|z_{ij}^+|}{\sum_{i}\left(|z_{ij}^+| + |z_{ij}^-|\right)}\\
\nu_{j}^{(l+1)} = R_{j}^{(l+1)} * \frac{\sum_{i}|z_{ij}^-|}{\sum_{i}\left(|z_{ij}^+| + |z_{ij}^-|\right)}
% \end{cases}\\
\\
\bar R_{i\in\mathcal{P,N}}^{(l)} = 
\sum_{j}
\Bigl(\frac{z_{ij}^+}{\sum_{i}\left(z_{ij}^+\right)} R_{j}^{(l+1)} + \frac{z_{ij}^{-}}{\sum_{i}\left(z_{ij}^{-}\right)}\nu_{j}^{(l+1)}
\Bigr)\\
\end{gathered}
\end{equation}
After propagating the relevance to the negatively contributed neurons, all neurons in layer \(R_{j}\) receive the relevance value, according to the inner ratio of each contribution. However, relevance is not conserved, and there is an over-allocation in layer $l$ in comparison to layer $l+1$. This over-allocated relevance, which originally means the negative contribution, is utilized for separating the important (unimportant) neurons by uniformly shifting all activated neurons. Let \(\Gamma=|\{m^{(l)}_{i}\neq 0\}|\) be the number of the activated neurons in layer \(i\). We evenly subtract the mean over allocation over these neurons. This shifts some of the relevancy scores to the negative region. Specifically, the shift is given by

\begin{equation}
\Psi_{i}^{l} = 
\begin{cases}
\sum_{i} \left(\bar R_{i \in \mathcal{N}}^{(l)}\right) * \frac{1}{\Gamma}  &,    \text{ $m_{i}^{(l)}$ is activated} \\
0 &,    \text{ otherwise} 
\end{cases}
\end{equation}

In this case, the relevance of the neurons in both groups $i\in \mathcal{P}$ and $i\in \mathcal{N}$ becomes
\begin{equation}
R_{i}^{(l)} = \bar R_{i\in \mathcal{P} \cup \mathcal{N}}^{(l)}-\Psi_{i}^{l}
\end{equation}

\begin{figure}
  \centering
  \includegraphics[scale=0.4]{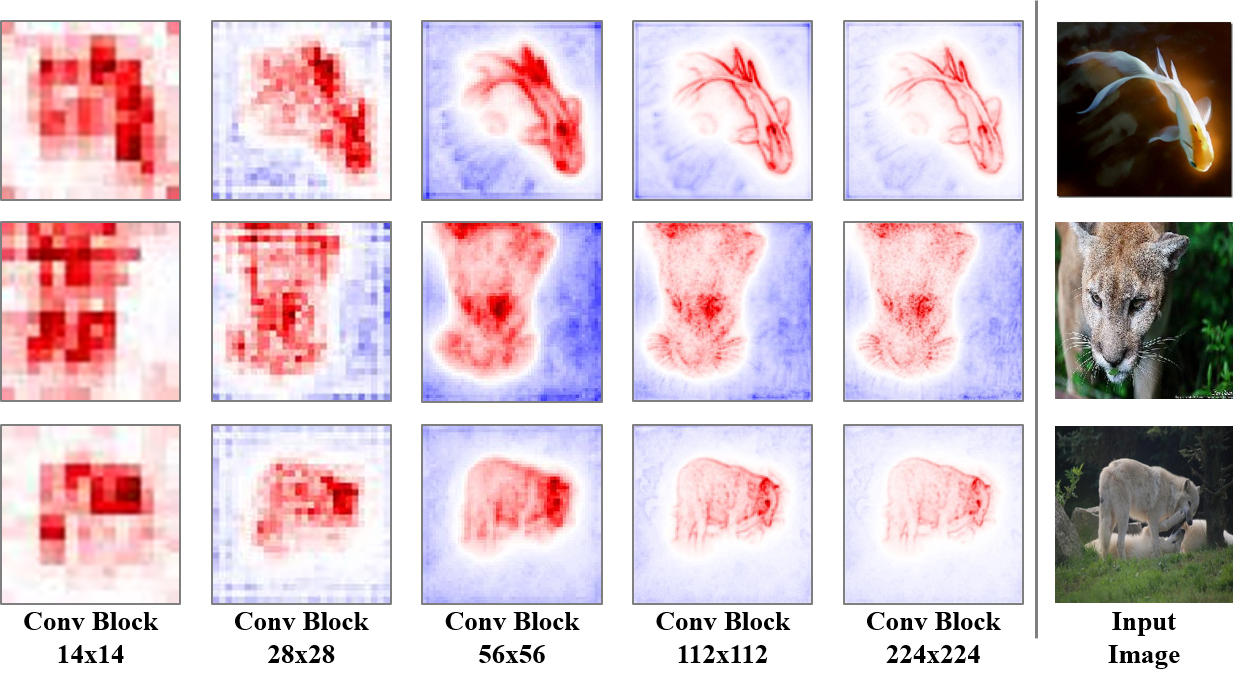}
  \caption{The visualization of the relevance map of the intermediate layers of the VGG-16 network.}
\end{figure}
From Eq.~8,9, it is easy to verify that the relevance value is preserved as in Eq.~2. We emphasize that the criterion for separating the important and unimportant neurons is the amount of each contribution, not the direction.
Since the activated parts are different between the feature maps of intermediate layers, we evenly subtract the same value to all activated neurons for not losing the important parts of each feature map. The negative input relevance for the next propagation indicates a relatively lower priority for the prediction result. Therefore, Eq.~7 propagate the negative relevance to the connected neurons, which contributed to the relatively unimportant neurons in positive and negative directions.

For the final relevance propagation to the input image layer, we utilized the \(Z^{\beta}\) rule \cite{bach2015pixel} which is commonly used for propagating to the input layer in methods derived from LRP. 
\begin{equation}
R_{i}^{(0)} = 
\sum_{j}
\Bigl(\frac{x_{i}w_{ij} - l_{i}w_{ij}^+ - h_{i}w_{ij}^-}
{\sum_{i}\left(x_{i}w_{ij} - l_{i}w_{ij}^+ - h_{i}w_{ij}^- \right)}R_{j}^{(1)}\Bigr)
\end{equation}
where \(x\) is the input image and \(\{l, h\}\) denotes the minimum and maximum values of \(x\).

We investigate the variation of the relevance maps during the propagating process. Fig. 4 shows the relevance map of activated neurons in the intermediate layers of VGG-16 net, where each pixel represents the sum of the relevance scores along the channel-wise axis. Here, we note that the intensity of the represented becomes lighter, since the size of the feature map is increased when passing the max-pooling layers, resulting in scattered to more neurons. As expected, the positive/negative maps change gradually from the classification layer (left) toward input ones (right).

\begin{figure*}[!t]
  \centering
  \includegraphics[width=\textwidth,scale=0.4]{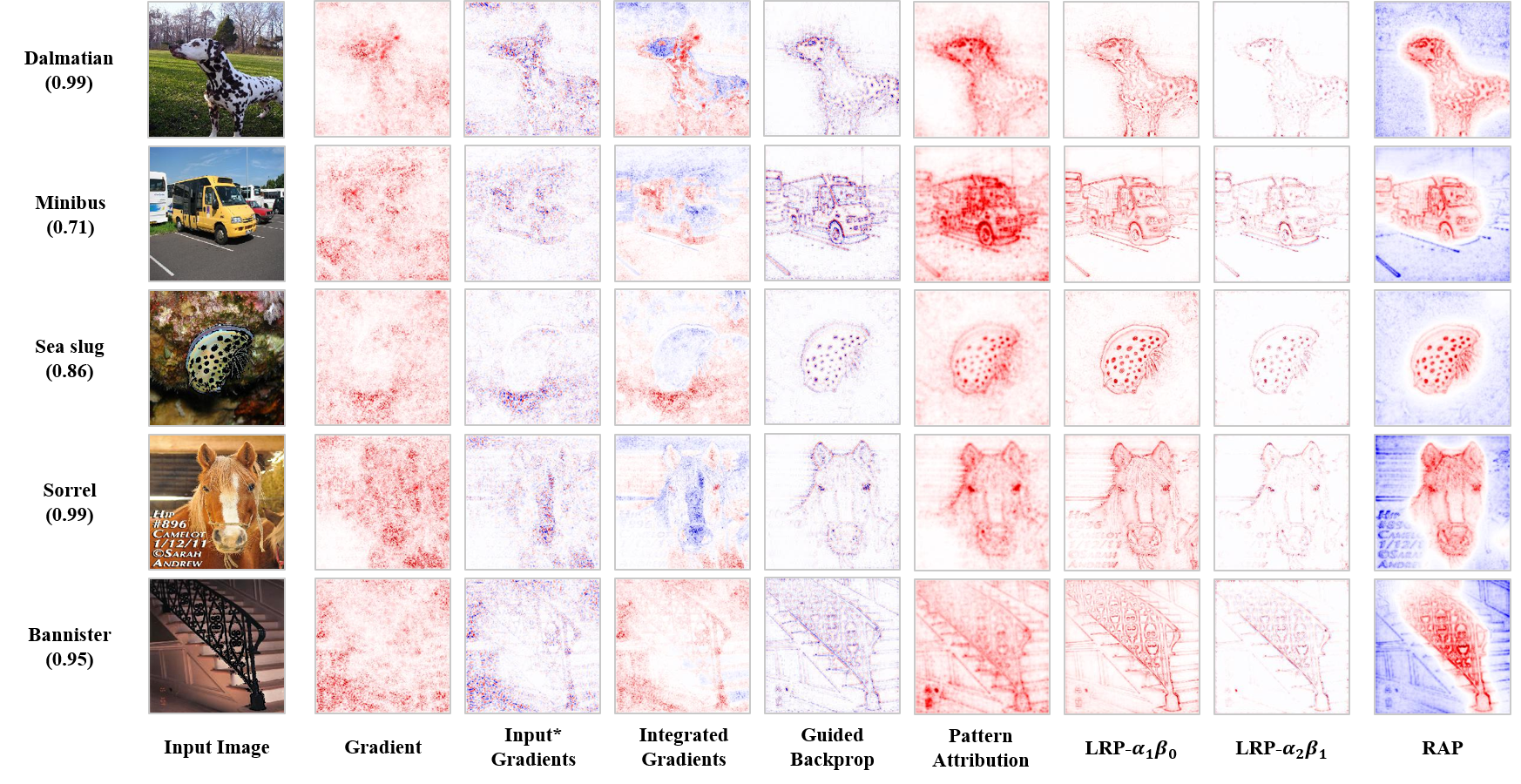}
  \caption{Comparison of the results of conventional methods and our RAP in VGG-16 network.}
\end{figure*}

\section{Experimental Evaluations}
We extensively verified our method on large scale CNNs, including the popular VGG-16 and ResNet-50 architectures. We utilize the Large Scale Visual Recognition Challenge 2012 (ILSVRC 2012) dataset \cite{ILSVRC15} and Pascal VOC 2012 dataset \cite{Everingham15}, which are widely employed and easily accessible. We also use the Imagenet segmentation mask provided by \cite{Guillaumin2014ImageNetAW}. We implement RAP with both Pytorch and Keras and visualize the explanation as a heatmap. For the evaluation, we utilized the Keras version to fairly compare with other explaining methods. We utilize the implementation introduced in \cite{alber2019innvestigate} for the other explaining methods. As is customary in the field (but maybe unintuitive), the visualized heatmap is represented by seismic colors, where red and blue colors denote positive and negative values, respectively.

The results of our method are compared with those of existing attribution methods, including integrated gradients, gradient, input* gradient, Guided BackProp, pattern attribution and LRP-\(\alpha\beta\) with \{\(\alpha_{1}\beta_{0}\), \(\alpha_{2}\beta_{1}\)\}. 
Since \cite{ancona2018towards} proved that LRP-\(\epsilon\) and DeepLift (Rescale version) are equivalent to the input* gradient when the model utilized the ReLU activation function, the result of input* gradient is the same as both methods. Since Pattern Attribution is developed for the VGG network, we do not report the result of it for experiments with Resnet or on the Pascal VOC dataset.
\subsection{Qualitative Evaluation of Heatmap}
For qualitatively evaluating the positive attributions generated by RAP, we compare the results by examining how the areas in which positive attributions converge are similar to those of the other methods. As the existing methods propagate the positive relevance well, we can utilize them to assess whether our method is consistent in attributing positive relevance. Fig. 5 presents the heatmaps generated from the various methods for the predicted images by the VGG-16 network. Fig. 6 illustrates the comparison between LRP-\(\alpha_{1}\beta_{0}\) and RAP in ResNet-50. More qualitative comparisons are illustrated in the supplementary material.

To qualitatively evaluate the negative attributions, we regard the attributions allocated in the parts that are not related to the prediction as to the negative relevance. While our results clearly distinguish between the object and the irrelevant parts as positive and negative attributions, the attributions from other methods overlap each other and appear as purple, as shown in Fig. 5. The results shown are typical: we qualitatively assessed all images in the validation set of ILSVRC 2012 and Pascal VOC 2012 dataset, and most of them appear to show similarly satisfactory results.
\begin{figure}[!h]
\centering
\includegraphics[scale=0.6]{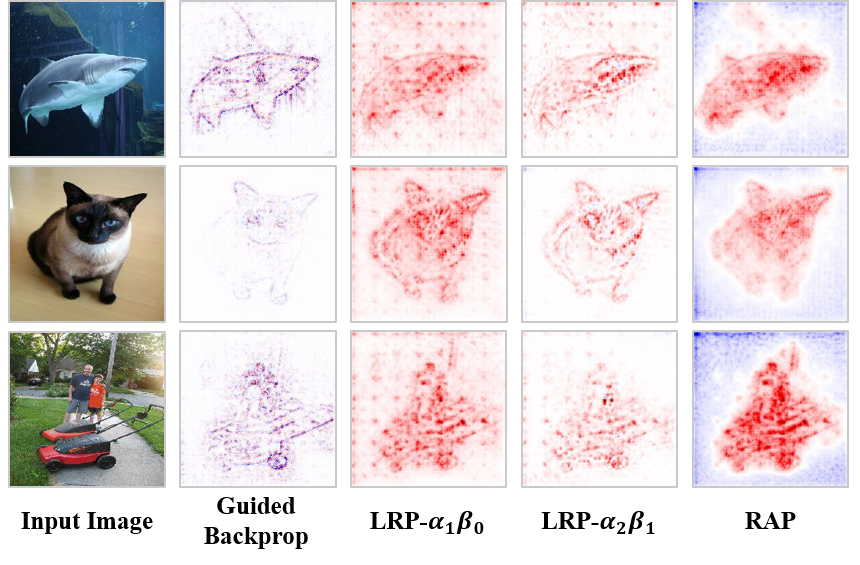}
\caption{Comparison of visualization results applied on ResNet-50.}
\end{figure}

\begin{table*}[t]
\centering
%\resizebox{\linewidth}{!}{
\begin{tabular}{c|c|c|c|c|c|c|c|c|c}
\toprule[0.5ex]
\multicolumn{2}{c|}{\multirow{2}{*}{Outside-Inside Ratio}} & \multirow{2}{*}{\textbf{RAP}} & \multirow{2}{*}{$LRP_{\alpha_{1}\beta_{0}}$} & \multirow{2}{*}{$LRP_{\alpha_{2}\beta_{1}}$} & \multirow{2}{*}{Gradient} & Input* & Integrated & Pattern & Guided\\
\multicolumn{2}{c|}{} &  &  &  &  & Gradient & Gradients & Attribution & Backprop\\\midrule[0.3ex]
\begin{tabular}[c]{@{}c@{}}VGG-16 \end{tabular} &
\begin{tabular}[c]{@{}c@{}}ALL\\POS \end{tabular} &
\begin{tabular}[c]{@{}c@{}}    \textbf{0.252}\\ \textbf{0.341}   \end{tabular} &
\begin{tabular}[c]{@{}c@{}}    -\\ 0.474   \end{tabular} &
\begin{tabular}[c]{@{}c@{}}    0.616\\ 0.524   \end{tabular} &
\begin{tabular}[c]{@{}c@{}}    -\\ 0.619   \end{tabular} &
\begin{tabular}[c]{@{}c@{}}    0.989\\ 0.691   \end{tabular} &
\begin{tabular}[c]{@{}c@{}}    1.230\\ 0.827   \end{tabular} &
\begin{tabular}[c]{@{}c@{}}    -\\ 0.415   \end{tabular} &
\begin{tabular}[c]{@{}c@{}}    1.069\\ 0.427   \end{tabular} \\\hline

\begin{tabular}[c]{@{}c@{}}Res-50 \end{tabular} &
\begin{tabular}[c]{@{}c@{}}ALL\\POS \end{tabular} &
\begin{tabular}[c]{@{}c@{}}    \textbf{0.164}\\ \textbf{0.166}   \end{tabular} &
\begin{tabular}[c]{@{}c@{}}    -\\ 0.429   \end{tabular} &
\begin{tabular}[c]{@{}c@{}}    0.302\\ 0.299   \end{tabular} &
\begin{tabular}[c]{@{}c@{}}    -\\ 0.597   \end{tabular} &
\begin{tabular}[c]{@{}c@{}}    0.996\\ 0.689   \end{tabular} &
\begin{tabular}[c]{@{}c@{}}    1.195\\ 0.698   \end{tabular} &
\begin{tabular}[c]{@{}c@{}}    -\\ -   \end{tabular} &
\begin{tabular}[c]{@{}c@{}}    1.035\\ 0.296   \end{tabular}\\
\midrule[0.3ex]
\multicolumn{2}{c|}{\multirow{2}{*}{Segmentation Mask}} & \multirow{2}{*}{\textbf{RAP}} & \multirow{2}{*}{$LRP_{\alpha_{1}\beta_{0}}$} & \multirow{2}{*}{$LRP_{\alpha_{2}\beta_{1}}$} & \multirow{2}{*}{Gradient} & Input* & Integrated & Pattern & Guided\\
\multicolumn{2}{c|}{} &  &  &  &  & Gradient & Gradients & Attribution & Backprop\\\midrule[0.3ex]
\begin{tabular}[c]{@{}c@{}}Imagenet \end{tabular} &
\begin{tabular}[c]{@{}c@{}}PIX ACC\\mIOU \end{tabular} &
\begin{tabular}[c]{@{}c@{}}    \textbf{79.23}\\ \textbf{62.23}   \end{tabular} &
\begin{tabular}[c]{@{}c@{}}    75.40\\ 55.78   \end{tabular} &
\begin{tabular}[c]{@{}c@{}}    72.95\\ 50.86   \end{tabular} &
\begin{tabular}[c]{@{}c@{}}    70.01\\ 49.30   \end{tabular} &
\begin{tabular}[c]{@{}c@{}}    66.38\\ 44.01   \end{tabular} &
\begin{tabular}[c]{@{}c@{}}    66,52\\ 45.90   \end{tabular} &
\begin{tabular}[c]{@{}c@{}}    76.84\\ 58.05   \end{tabular} &
\begin{tabular}[c]{@{}c@{}}    71.98\\ 49.87   \end{tabular} \\\hline

\begin{tabular}[c]{@{}c@{}}Pascal VOC \end{tabular} &
\begin{tabular}[c]{@{}c@{}}PIX ACC\\mIOU \end{tabular} &
\begin{tabular}[c]{@{}c@{}}    \textbf{73.91}\\ \textbf{55.60}   \end{tabular} &
\begin{tabular}[c]{@{}c@{}}    70.86\\ 49.82   \end{tabular} &
\begin{tabular}[c]{@{}c@{}}    69.43\\ 46.85   \end{tabular} &
\begin{tabular}[c]{@{}c@{}}    68.14\\ 46.07   \end{tabular} &
\begin{tabular}[c]{@{}c@{}}    50.01\\ 31.69   \end{tabular} &
\begin{tabular}[c]{@{}c@{}}    52.38\\ 34.39   \end{tabular} &
\begin{tabular}[c]{@{}c@{}}    -\\ -   \end{tabular} &
\begin{tabular}[c]{@{}c@{}}    66.92\\ 43.63   \end{tabular}
\\
\bottomrule[0.5ex]
\end{tabular}%}
\caption{In Outside-inside relevance ratio result, the first (second) row is the ratio when considering all (only positive) relevance in Imagenet dataset. Segmentation mask result shows the pixel accuracy and mIOU of relevance heatmap in VGG-16 network.}
\label{tab:1}
\end{table*}
\begin{table}[]
\begin{small}
    \centering
    \begin{tabular}{@{}l|c@{}}
        \toprule[0.5ex]
        %\multicolumn{2}{c}{ImageNet-Segmentation dataset}\\
        Method & mIOU\\
        \midrule
         Guillaumin et al.~\cite{guillaumin2014imagenet} & 57.30\\
         DeepMask~\cite{pinheiro2015learning} &  58.69\\
         DeepSaliency~\cite{li2016deepsaliency} &  62.12\\
         Xiong et al.~\cite{xiong2018pixel} & 64.22\\
        \midrule
         Ours & 62.23\\
         \bottomrule[0.5ex]
    \end{tabular}
    \end{small}
    \caption{Quantitative mIOU results on the ImageNet Segmentation task. Our method is highly comparable to state of the art, despite not using the additional supervision.}
    \label{tab:imagenet_seg}
\end{table}
\subsection{Quantitative Assessment of Attributions}
It is not trivial to assess the quantitative performance of the methods designed for explaining DNN models, since each method poses a different assumption and is designed for slightly different objectives. In this work, we utilized three methods that are commonly used to evaluate objectness and relevance: (i) Outside-Inside ratio, (ii) Pixel accuracy and Intersection of Union, and (iii) Region perturbation. 

\cite{lapuschkin2016analyzing} introduce a method for assessing how the attributions are focused on the object, by computing the relevance inside and outside of the bounding box. We extend this method to consider the effect of correctly and wrongly distributed negative relevance. However, since the bounding box is not perfect for the object corresponding to the prediction, we additionally utilize the segmentation masks and metrics to assess how the positive attributions are correctly distributed in the target object. \cite{samek2017evaluating} introduce the method for quantitatively assessing the explanations methods, which utilizes the region perturbation process that progressively distorts the pixels from the heatmap and formalized this method as Area over the perturbation curve (AOPC). To evaluate how the negative attributions are distributed to the irrelevant regions of the prediction, we perturbate \textit{least relevant first} (LeRF) and investigate the degradation of the accuracy.

In our experiment, we extract 10,000 correctly classified images from the validation set of imagenet and employ the specified bounding boxes (for the pertubation test). For the objectness scores, we utilized the 4,276 images from the imagenet dataset with segmentation masks and 1,449 images from the Pascal VOC validation set.
\subsubsection{Objectness of Positive Attributions}
To verify how the attributions are distributed on the prediction object, we assess the outside-inside relevance ratio of attributions~\cite{lapuschkin2016analyzing}, utilizing the bounding box. We extend the original metric to evaluate the positive and negative relevance simultaneously.
\begin{equation}
    \mu = \frac{\frac{1}{\left\vert P_{out} \right\vert}\sum_{q\in P_{out}}R_q^{(0)+}+\frac{1}{\left\vert P_{in} \right\vert}\sum_{p\in P_{in}}R_p^{(0)-}}
    {\frac{1}{\left\vert P_{in} \right\vert}\sum_{p\in P_{in}}R_p^{(0)+} + \frac{1}{\left\vert P_{out} \right\vert}\sum_{q\in P_{out}}R_q^{(0)-}}
\end{equation}
Here, \(\left\vert \cdot \right\vert\) is the cardinality operator and \(P_{in,out}\) denotes the set of pixels inside and outside of the bounding box, respectively. When the positive (negative) relevance is attributed out of the bounding box, the value of \(\mu\) is increased (decreased). By contrast, if the positive (negative) relevance is distributed in the bounding box, the ratio is decreased (increased). For the conventional methods which only consider the positive relevance, this metric becomes identical to the original metric of \cite{lapuschkin2016analyzing}. 

We present the results with(out) considering the negative relevance. In Tab.~\ref{tab:1}, this is denoted as ALL when considering both and POS when discarding the negative relevance. For both cases in Tab.~\ref{tab:1}, RAP provides the best scores with indicating that the relevance attribution is better distributed inside/outside the bounding box than the existing methods.

\begin{figure*}
\centering
\includegraphics[width=.90\linewidth]{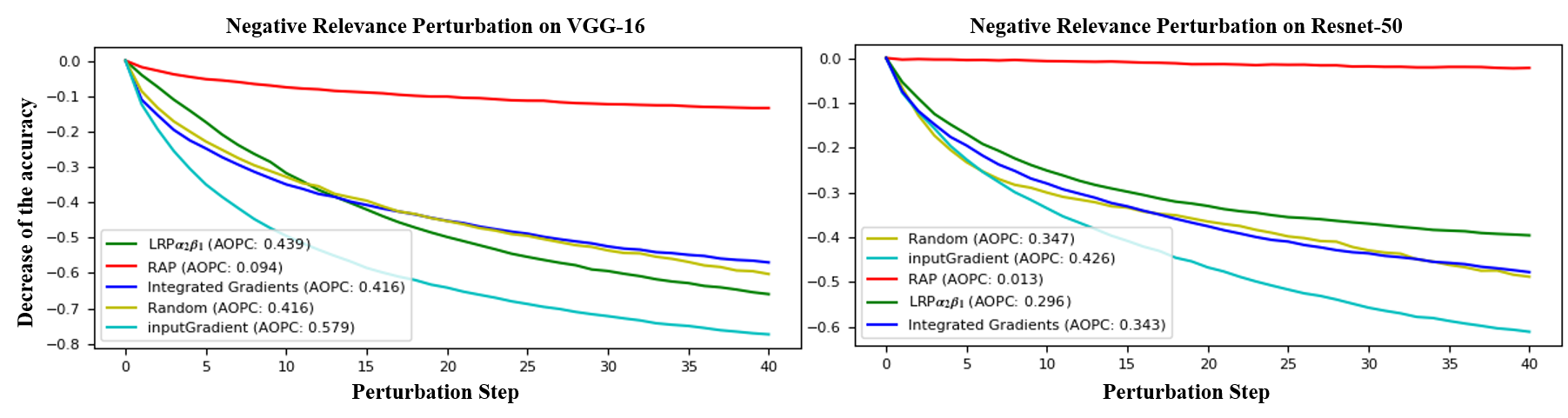}
\caption{This graph illustrates the results of the negative perturbation on VGG-16 and Resnet-50. For each step, 100 pixels corresponding to the LeRF is perturbated as zero. RAP shows the unique characteristics of the robustness to the perturbation.}
\label{fig:was7}
\end{figure*}

Furthermore, we utilize the segmentation masks provided for the ImageNet and Pascal VOC 2012 validation sets to evaluate our explainability performance in terms of objectness. The evaluation is done by first producing an explainability map, followed by thresholding, considering two cases: (i) only positive relevance, in which case the mean value is taken as a threshold, and (ii) positive and negative relevance, where the threshold is set to zero. By thresholding, we produce a segmentation mask, considering values above the threshold as 1, and 0 otherwise.
Tab. 1 shows that our method greatly outperforms previous work in both Pixel Accuracy and mean-IOU. These results are stable for other choices of the threshold as well, such as various percentiles and using the median. As can be seen, the output relevance is highly correlated with the input image objectness.

Interestingly, as shown in Tab.~\ref{tab:imagenet_seg}, our method is extremely competitive with respect to the objectness literature methods, which are trained with additional supervision, such as bounding boxes. This is demonstrated using the acceptable metric of mIOU on imagenet, which is commonly used for benchmarking the generic object segmentation \cite{cho2017novel}.
\subsubsection{Interfering Negative Attributions}
When a DNN makes a correct prediction, removing the contribution of irrelevant pixels from the prediction should not change significantly the prediction accuracy and relevance values. A small decrement of accuracy is inevitable because the distortion of color and shape could affect the classification, resulting in unpredictable noise and attack. We carefully discuss these issues in more details in supplementary materials. Also, it is important to note that removing pixels corresponding to the negative attributions does not always bring an increment of the accuracy, because the negative relevance of incorrect prediction does not denote the positive relevance of true label. 

Fig.~\ref{fig:was7} shows the results when applying the LeRF perturbation on the VGG-16 and Resnet-50. For each step, we perturbate 100 pixels corresponding to the negative relevance, a total of 4,000 pixels distortion. As shown in the results, while other methods that consider the negative relevance show the rapid decrements of the accuracy, RAP rarely affects the prediction result during the negative attributions removal process. Thus, RAP distinguishes between the relevant and unimportant parts of the input image without overlapping each other. 
\section{Conclusion}
In this paper, we propose RAP, a new method for interpreting the neurons in terms of importance to the predictions of DNNs, by assigning the relevance score according to the influence of each neuron. By approaching the relevance in terms of influence among the neurons, it is possible to separate relevant and irrelevant regions to the prediction. We evaluate our methods in quantitative and qualitative ways to verify that the attributions correctly account for the meaning. For the quantitative evaluation, we utilize the metrics: Outside-Inside ratio, mIOU and region perturbation to confirm how the attributions focused on the (ir)relevant object according to their assigned relevance scores. Overall, the experiments show that the RAP method leads to the desired characteristics: (i) a clear distinction of positive (relevant) and negative (irrelevant) attributions and (ii) it is indicative of objectness and can separate the main image object from the other irrelevant regions.
\section{Acknowledgments}
This work was supported by Institute for Information \& communications Technology Planning \& Evaluation(IITP) grant funded by the Korea government(MSIT) (No.2017-0-01779, A machine learning and statistical inference framework for explainable artificial intelligence \& No.2019-0-01371, Development of brain-inspired AI with human-like intelligence) and the European Research Council (ERC) under the European Unions Horizon 2020 research and innovation programme (grant ERC CoG 725974).

{\small
\bibliographystyle{aaai}
\bibliography{egbib}
}
\end{document}